\pdfoutput=1

\documentclass[11pt]{article}

\usepackage{acl}

\usepackage{times}
\usepackage{latexsym}

\usepackage[T1]{fontenc}

\usepackage[utf8]{inputenc}

\usepackage{microtype}

\usepackage{inconsolata}

\usepackage{graphicx}
\usepackage{multirow}
\usepackage{tabularx} 
\usepackage{booktabs} 
\usepackage{amsmath}
\usepackage{float}

\title{Dovetail: A CPU/GPU Heterogeneous Speculative Decoding for LLM inference}

\author{
 \textbf{Libo Zhang\thanks{\ \ indicates equal contribution.}},
 \textbf{Zhaoning Zhang$^*$\thanks{\ \ indicates corresponding authors.}},
 \textbf{Baizhou Xu},
 \textbf{Rui Li},
 \textbf{Zhiliang Tian}, \\
 \textbf{Songzhu Mei},
 \textbf{Dongsheng Li}
 \\
 National Key Laboratory of Parallel and Distributed Computing \\ 
 College of Computer Science and Technology \\
 National University of Defense Technology, Changsha, China.
 \\
 \texttt{ \{zhanglibo, zhangzhaoning, xubaizhou23,lirui.r21,} \\
 \texttt{tianzhiliang, sz.mei, dsli\}@nudt.edu.cn} }

\begin{document}
\maketitle
\begin{abstract}

With the continuous advancement in the performance of large language models (LLMs), their demand for computational resources and memory has significantly increased, which poses major challenges for efficient inference on consumer-grade devices and legacy servers. These devices typically feature relatively weaker GPUs and stronger CPUs. Although techniques such as parameter offloading and partial offloading can alleviate GPU memory pressure to some extent, their effectiveness is limited due to communication latency and suboptimal hardware resource utilization. To address this issue, we propose Dovetail\footnote{\url{https://github.com/ddInference/Dovetail}}, a lossless inference acceleration method that leverages the complementary characteristics of heterogeneous devices and the advantages of speculative decoding. Dovetail deploys a draft model on the GPU to perform preliminary predictions, while a target model running on the CPU validates these outputs. By reducing the granularity of data transfer, Dovetail significantly minimizes communication overhead. To further improve efficiency, we optimize the draft model specifically for heterogeneous hardware environments by reducing the number of draft tokens to lower parallel verification latency, increasing model depth to enhance predictive capabilities, and introducing a Dynamic Gating Fusion (DGF) mechanism to improve the integration of feature and embedding information. We conduct comprehensive evaluations of Dovetail across various consumer-grade GPUs, covering multiple tasks and mainstream models. Experimental results on 13B models demonstrate that Dovetail achieves inference speedups ranging from 1.79× to 10.1× across different devices, while maintaining consistency and stability in the distribution of generated texts.


\end{abstract}

\section{Introduction}

In recent years, with the continuous growth of model parameter scales, large language models (LLMs) \cite{touvron2023llama,achiam2023gpt} have achieved significant performance improvements across multiple domains. However, their substantial computational and memory demands impose higher requirements on hardware \cite{tang2024delta}, posing severe challenges for deployment on personal or consumer-grade devices, including outdated servers from the pre-large-model era.
\begin{figure}[t]
  \includegraphics[width=\columnwidth]{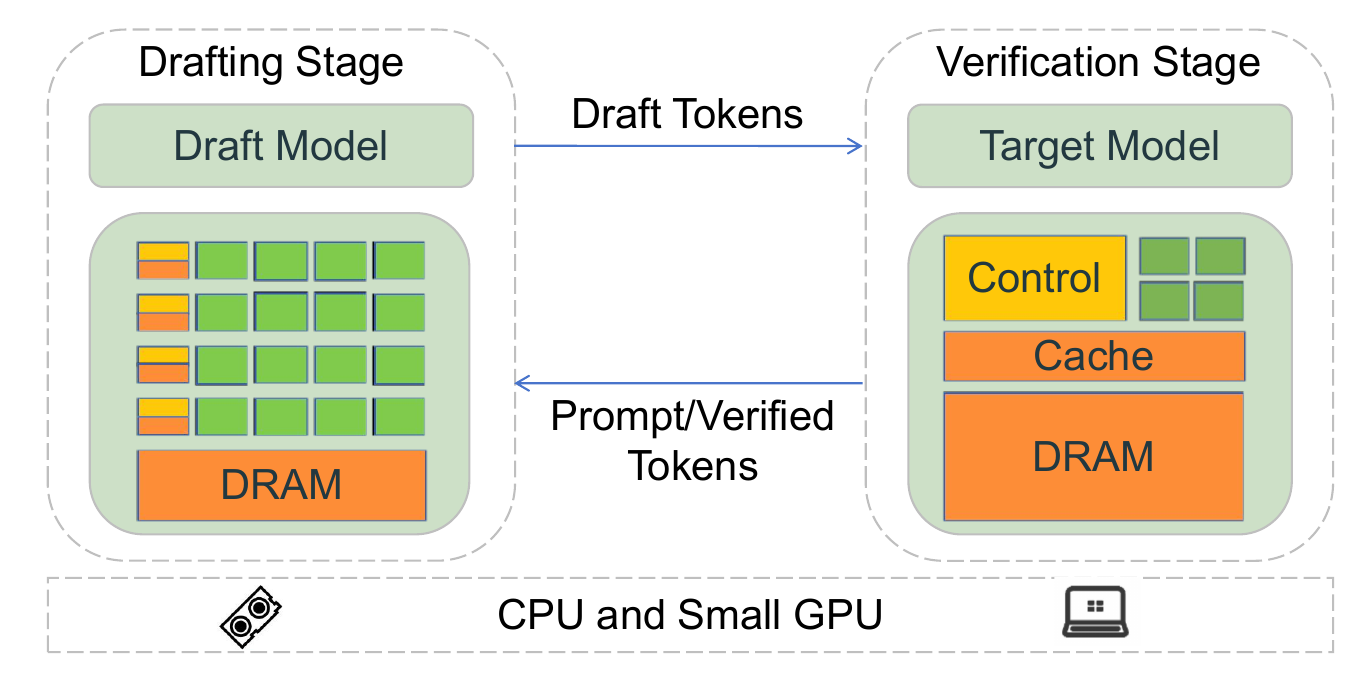}
  \caption{The architecture of Dovetail, highlighting a collaborative inference mode where the target model is deployed on the CPU, and the draft model is deployed on the GPU. }
  \label{fig:architecture}
\end{figure}


\begin{figure*}[t]
  \includegraphics[width=\linewidth]{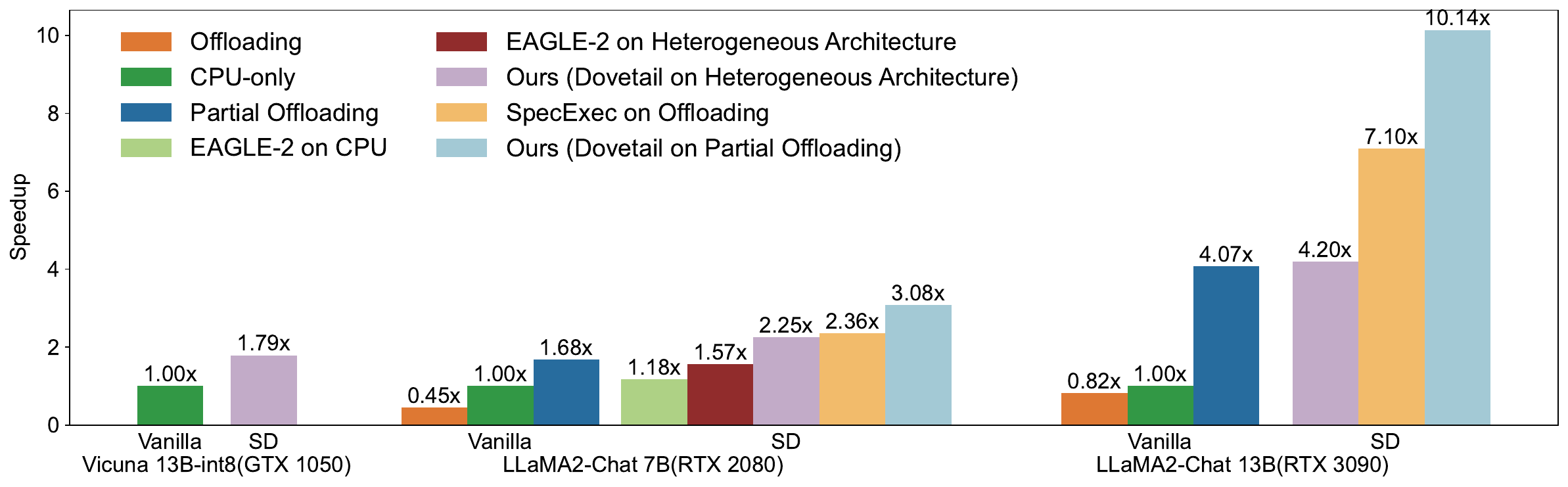}
  \caption{The speedup ratios of different models were tested on consumer-grade devices with temperature = 0. "Vanilla" refers to the existing lossless inference method, while "SD" stands for speculative decoding, including the effects when applying the SD algorithm on top of the Vanilla method.}
  \label{fig:compare-ratios}
\end{figure*}

We observe that these devices and small-scale servers are typically equipped with GPUs with limited memory, making it difficult to fully load LLMs. For instance, a 7B model requires approximately 14GB of memory at 16-bit precision, far exceeding the capacity of consumer-grade GPUs such as the NVIDIA RTX 2080. Currently, the primary strategies for conducting inference without compromising model performance are offloading and partial offloading. The former temporarily stores a portion of parameters in host memory and dynamically loads them into the GPU, while the latter directly executes part of the computation on the CPU, thereby alleviating memory pressure. As shown in Figure~\ref{fig:compare-ratios}, compared to pure CPU inference, offloading\cite{DBLP:conf/nips/SvirschevskiMCC24} reduces inference speed to 0.45x the original due to high communication latency between the CPU and GPU. Partial offloading improves this to 1.68x, but the acceleration effect is constrained by GPU memory capacity, diminishing as available memory decreases.

Speculative Decoding \cite{DBLP:conf/icml/LeviathanKM23,chen2023accelerating} is an emerging method for accelerating LLM inference. It leverages a smaller model to generate multiple draft tokens, which are then verified in parallel by the target model, enabling the generation of multiple tokens in a single forward pass without losing performance. Although SpecExec \cite{DBLP:conf/nips/SvirschevskiMCC24} applies this technique to offloading scenarios to accelerate inference, it still suffers from high communication latency, inefficient utilization of hardware resources, and requires at least 5.9 GB of GPU memory in the current test environment, making it difficult to deploy effectively on devices with lower memory. To address these issues, we propose Dovetail, a heterogeneous CPU-GPU collaborative speculative decoding mechanism, as illustrated in Figure~\ref{fig:architecture}. In this setup, the draft model is deployed on a consumer-grade GPU, while the target model executes on the CPU. By reducing the granularity of data transfer from Transformer blocks to tokens, Dovetail significantly reduces communication overhead. Additionally, thanks to the flexible parameter scale of the draft model (ranging from 68M to 3B), Dovetail can operate efficiently on most consumer-grade GPUs.

As shown in Figure~\ref{fig:compare-ratios}, when directly applying speculative decoding algorithms on heterogeneous architectures, the acceleration effect is only improved by 1.57 times. To further enhance inference speed on such architectures, we explore the characteristics of speculative decoding algorithms in this context and optimize the existing approach as follows: By reducing the number of candidate draft tokens, we linearly decrease the latency of parallel verification, effectively mitigating performance bottlenecks on low-end hardware. Given the significant increase in target model latency, adopting a larger draft model becomes feasible. Based on EAGLE-2 \cite{DBLP:conf/emnlp/LiW0024}, we redesign the draft model by introducing DGF to dynamically adjust the fusion weights between hidden states and token embeddings, avoiding information loss and imbalance in feature representation fusion. Furthermore, by expanding the draft model’s Transformer blocks from single to multiple, we significantly narrow the performance gap between the draft and target models while improving prediction performance and increasing the average acceptance length.

Our main contributions include:
\begin{enumerate}
\item We propose a novel heterogeneous speculative decoding paradigm that fully leverages the characteristics of heterogeneous architectures and speculative decoding. By deploying the target model’s verification phase on the CPU, this paradigm significantly improves hardware resource utilization efficiency.
\item We optimize the existing draft model for low-end hardware in heterogeneous architectures, achieving a better balance between latency and performance.
\item We develop a system that requires only 3GB of VRAM to achieve an inference speed of 4.62 to 5.86 tokens per second for models such as LLaMA2-Chat 7B, demonstrating a 2.25x performance improvement on MT-bench compared to existing methods. When the VRAM is increased to 7GB, the inference speed further improves to 6.5 to 8 tokens per second, resulting in a performance enhancement of 3.08x. On the GeForce RTX 3090, tests on LLaMA2-Chat 13B indicate that our method achieves a maximum speedup ratio of 10.14x.

\end{enumerate}

\section{Preliminaries}


\subsection{Effectiveness of Heterogeneous Speculative Decoding}  
In resource-constrained environments, computational resources typically consist of a combination of CPUs and small-scale GPUs, such as CPUs paired with discrete GPUs (dGPUs) or integrated GPUs (iGPUs) in personal devices, as well as CPUs paired with small-scale GPUs in servers. These configurations are not specifically designed for AI, and mainstream methods achieve large language model (LLM) inference through parameter offloading. Given the characteristics of computational resource configurations and the properties of speculative decoding, we propose a heterogeneous speculative decoding method to accelerate LLM inference. However, this method may not perform well in all combinations of main processors and accelerators. Therefore, we employ stochastic analysis to reveal the correlation between hardware and computational configurations. For a detailed analysis, please refer to Appendix~\ref{sec:appendix}.

\subsection{Factors Affecting Speculative Decoding Speedup}\label{sec:2.2}

The time for the target model to decode a single token is \( T_T \), while the time for the speculative decoding algorithm to decode a single token is \( T_{\mathit{Avg}}^{\mathit{SD}} \). The performance analysis formula \cite{DBLP:conf/iclr/SadhukhanCCTLSY25} can be expressed as:  
\begin{equation}  
\frac{T_{\mathit{Avg}}^{\mathit{SD}}}{T_T} = \frac{1}{\Omega(\gamma, \alpha)} \left( \frac{\gamma \cdot T_D}{T_T} + \frac{T_V(\gamma)}{T_T} \right)
\label{eq:speedup-breakdown}  
\end{equation}  
where \( \alpha \) is the acceptance rate, \( \gamma \) is the number of candidate draft tokens, \( \Omega(\gamma, \alpha) \) is the number of accepted tokens in a single parallel verification, \( T_D \) is the time for the draft model to decode a single token, and \( T_V(\gamma) \) is the time for the target model to verify \( \gamma \) tokens in parallel.  
\begin{figure}[H]
  \includegraphics[width=\columnwidth]{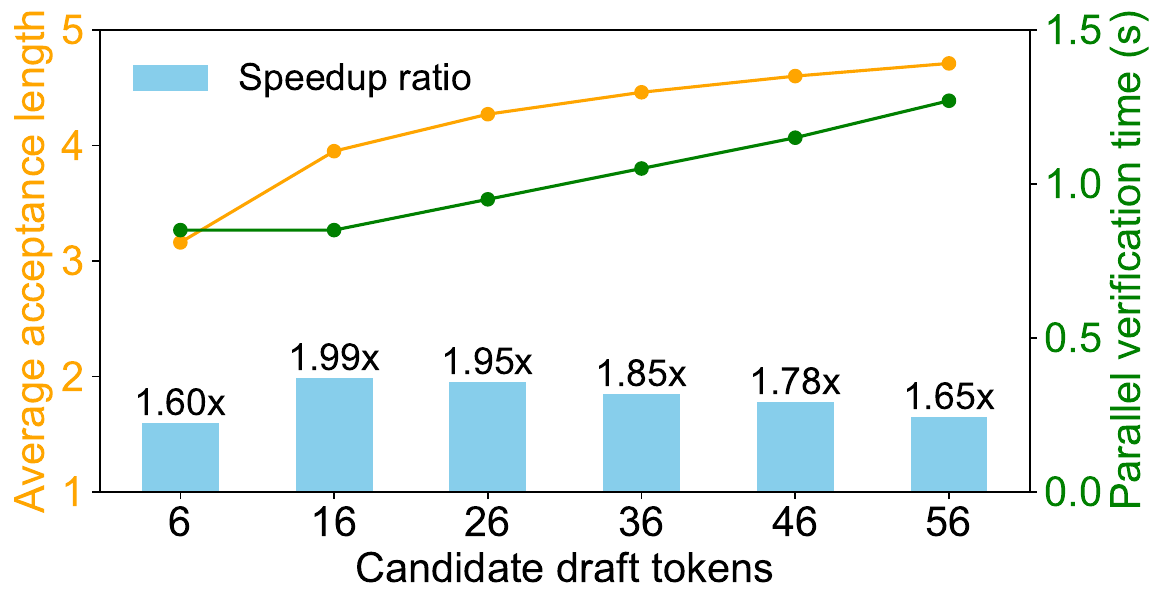}
  \caption{Explore the interrelationship between the average acceptance length \( \Omega(\gamma, \alpha) \), parallel validation time \( T_V(\gamma) \), and speedup ratio of the target model under different candidate draft tokens \( \gamma \).}
  \label{fig:2-13}
\end{figure}
\begin{table}[htbp]
\centering
\resizebox{\columnwidth}{!}{%
\begin{tabular}{ccccccccc}
\toprule
Method & Device & \( \gamma \) & $T_T$ & $T_D$ & \( T_V(\gamma) \) & \( \Omega(\gamma, \alpha) \) & Speedup \\
\midrule
EAGLE-2 & GPU 4090 & 60 & 0.025 & 0.001 & 0.025 & 5.59 & 4.01$\times$ \\
EAGLE-2 & GPU 2080 & 60 & 0.45 & 0.0016 & 1.34 & 5.61 & 1.90$\times$ \\
EAGLE-2 & GPU 2080 & 16 & 0.45 & 0.0016 & 0.88 & 4.69 & 2.32$\times$ \\
Dovetail & GPU 2080 & 16 & 0.45 & 0.0072 & 0.88 & 5.90 & 2.77$\times$ \\
\bottomrule
\end{tabular}%
}
\caption{Performance comparison of different methods and hardware configurations on the HumanEval dataset. $T_T$, $T_D$, and \( T_V(\gamma) \) denote the time costs of target model computation, draft model computation, and verification, respectively. For GPU 2080, the target model resides on the CPU while the draft model is on the GPU; for GPU 4090, both models are placed on the GPU.}
\label{tab:performance_comparison}
\end{table}

The key factors influencing the acceleration effect include: \( T_D / T_T \), \( T_V(\gamma) / T_T \), and \( \Omega(\gamma, \alpha) \). As shown in Table~\ref{tab:performance_comparison}, the results in the first and second rows demonstrate that in resource-constrained heterogeneous architectures, 
$T_D/T_T$ approaches zero while $\Omega(\gamma,\alpha)$ remains constant. However, due to the limited parallelism of CPUs \cite{yin2021parax}, 
$T_V(\gamma)/T_T$ increases significantly, leading to a degradation in overall acceleration performance.


The increase in \( T_V(\gamma) \) shifts the primary bottleneck of heterogeneous speculative decoding to the parallel verification process of the target model. Reducing the number of draft tokens can lower \( T_V(\gamma) \), but it also shortens \( \Omega(\gamma, \alpha) \). Therefore, a balance must be struck between the two. As shown in Figure~\ref{fig:2-13}, reducing the number of draft tokens linearly decreases verification latency. Although the average acceptance length is reduced, the overall inference speed still improves.

As \( T_V(\gamma) \) decreases and stabilizes, the primary bottleneck shifts to \( \Omega(\gamma, \alpha) \). Increasing \( \alpha \) is typically accompanied by an increase in \( T_D \). Research by DSD \cite{yan2024decoding} indicates that enlarging the parameter size of the draft model can enhance \( \Omega(\gamma, \alpha) \). However, the continuous rise in \( T_D \) causes the overall inference speed to first increase and then decrease. In heterogeneous architectures, the increase in \( T_T \) is much greater than that in \( T_D \), resulting in a significant reduction in \( T_D / T_T \). This allows for the deployment of draft models with larger parameter sizes, thereby increasing \( \alpha \), extending \( \Omega(\gamma, \alpha) \), and ultimately improving the overall inference speed.  

\begin{figure}[t]
  \centering
  \includegraphics[width=\columnwidth]{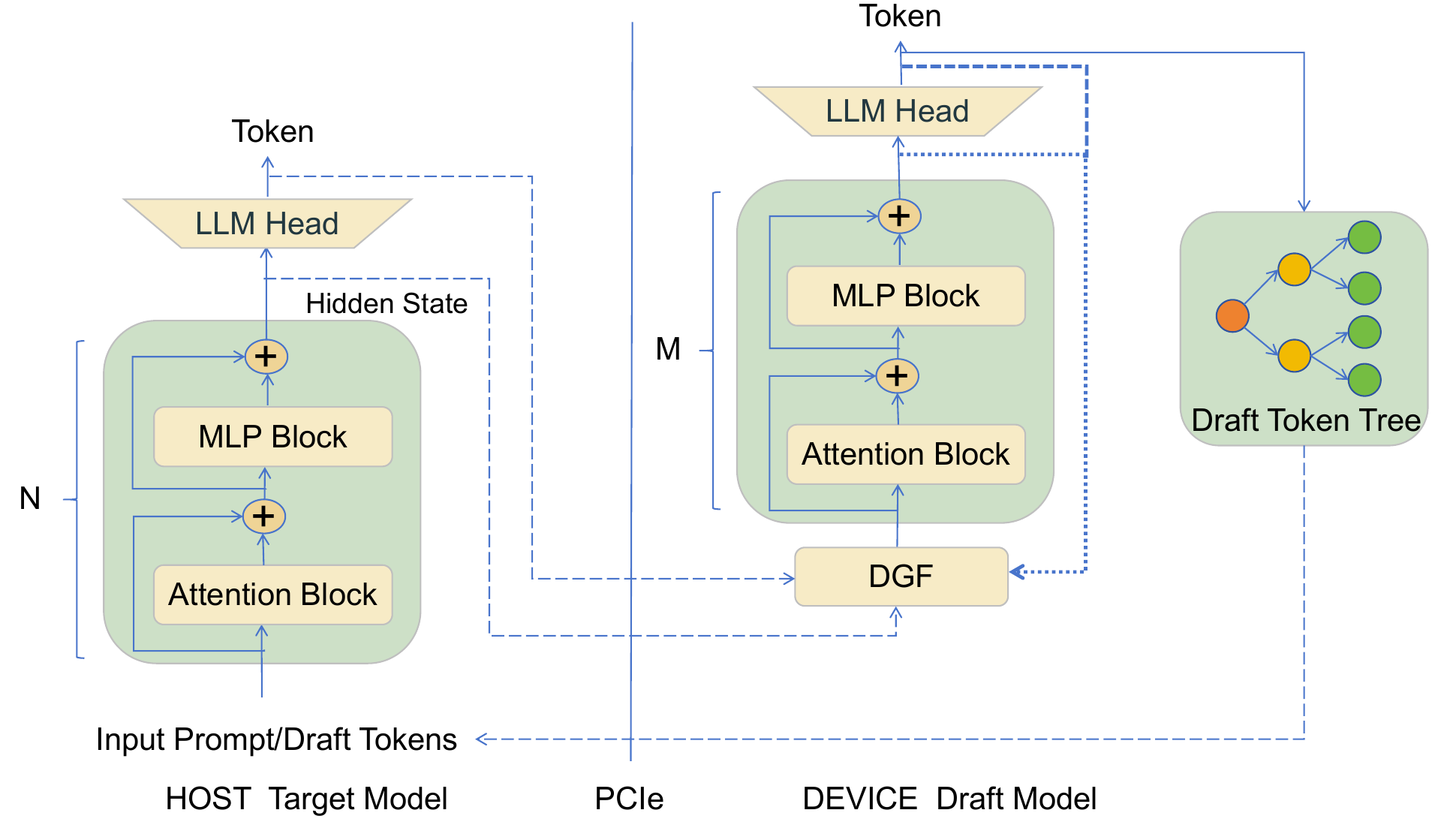}
  \caption{The pipeline of heterogeneous collaborative speculative decoding depicts the computational procedure. In this context, N and M denote the number of layers in the target model and the draft model, respectively.}
  \label{fig:4.1}
\end{figure}

Based on this, the key to optimizing the performance of heterogeneous speculative decoding lies in: linearly reducing \( T_V(\gamma) \) by decreasing \( \gamma \), while employing draft models with larger parameter sizes to increase \( \alpha \), thereby enhancing \( \Omega(\gamma, \alpha) \) and achieving overall performance optimization.


\section{Method}
In this section, we provide a detailed description of the implementation of Dovetail.

\subsection{CPU/GPU Heterogeneous Architecture}

Dovetail employs a CPU/GPU heterogeneous architecture, where the draft model is deployed on the GPU and the target model on the CPU, leveraging the advantages of heterogeneous computing. As illustrated in Figure~\ref{fig:4.1}, the target model first processes the input prompt to generate the hidden states required by the draft model. These states, along with the corresponding tokens, are transferred to the GPU for draft token generation.The draft model dynamically constructs a draft tree through multiple rounds of autoregressive decoding. In each round, it computes the cumulative product of token generation probabilities along the path from the root node to each leaf node, which is treated as the global acceptance probability. Based on these probabilities, the top-k tokens are selected for further decoding and expansion in the next round. This process repeats until the tree is fully constructed.Once constructed, the tree nodes are reordered according to their global acceptance probabilities, and the top-\(\gamma\) tokens with the highest scores are chosen as candidates. These candidates are then sent to the target model on the CPU for parallel verification. The target model computes the logits of the candidate tokens in a single forward pass and applies a speculative sampling algorithm to determine the accepted tokens. The accepted tokens are returned to the GPU-based draft model to initiate the next round of draft tree generation. 



\subsection{Dynamic Gated Fusion}

In EAGLE-2, the draft model requires the fusion of hidden states and token embeddings to address the uncertainty of hidden states before inference. The current method simply concatenates the two and maps them to the hidden state dimension through a single linear transformation. However, this approach has limitations: (1) it may cause the model to overly rely on linear transformations, neglecting the deep interaction between hidden states and token embeddings; (2) the fixed linear layer lacks flexibility when processing features from different levels, unable to dynamically adjust the fusion process based on context, which may lead to insufficient emphasis on critical information and affect fusion performance.
\begin{figure}[t]
  \centering
  \includegraphics[width=0.8\columnwidth]{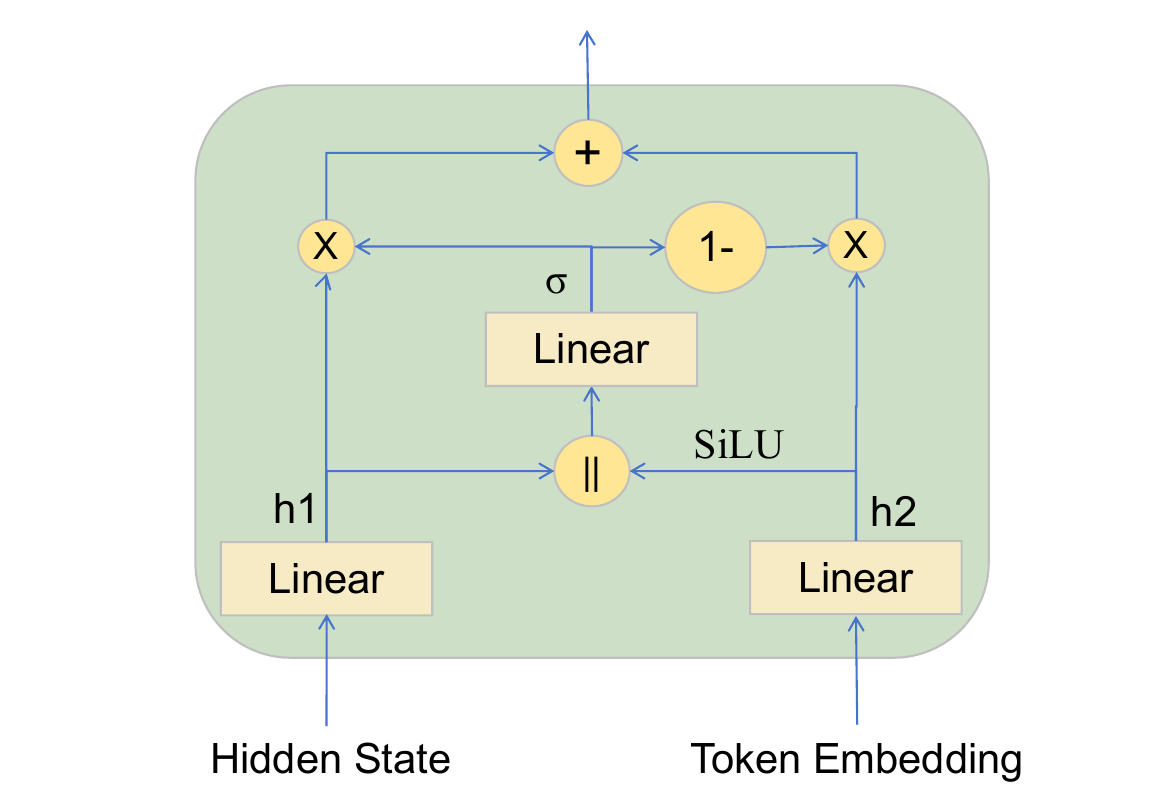}
  \caption{Schematic diagram of the DGF. Hidden State represents the second-to-top hidden state in the target LLM, “\textbar\textbar” denotes concatenation, “$\sigma$” represents the sigmoid function, and “x” denotes the multiplication mechanism.}
  \label{fig:4.22}
\end{figure}
To address these issues, we propose the Dynamic Gated Fusion (DGF) module, inspired by multimodal feature fusion \cite{DBLP:conf/iclr/OvalleSMG17}. As shown in Figure~\ref{fig:4.22}, the DGF module first applies linear transformations to the hidden states and token embeddings, generating feature representations \( h_1 \) and \( h_2 \), which are then concaenated into a joint feature vector. A linear layer and sigmoid activation function generate gating values to dynamically adjust the contribution ratios of \( h_1 \) and \( h_2 \), and a weighted sum produces the fused feature representation, effectively capturing their interaction. Compared to the method of concatenation followed by linear transformation, DGF adaptively regulates the interaction strength between hidden states and token embeddings and dynamically adjusts the fusion ratio based on input scenarios, enhancing the model’s expressive power in complex contexts while reducing the risk of information loss or fusion imbalance caused by global linear transformations.

\subsection{Multiple Transformer Blocks}

In high-performance hardware environments, the latency of the draft model is the primary bottleneck for the speedup ratio of speculative decoding algorithms. Therefore, designing the draft model requires balancing parameter scale and prediction accuracy. Typically, the draft model adopts a smaller parameter scale, such as a single Transformer block of the target model, to achieve significant inference acceleration. However, in resource-constrained heterogeneous architectures, this design often leads to insufficient performance. As discussed in Section \ref{sec:2.2}, the increased parallel verification time of the target model provides opportunities for optimizing the draft model. Although the Dynamic Gated Fusion (DGF) module can effectively integrate information from different layers to improve performance, its single-Transformer-block architecture limits its ability to learn deep abstract features of the target model and align feature distributions, constraining prediction accuracy.

Based on this, we propose extending the draft model to a multi-block architecture with \( M \) Transformer blocks, as shown in Figure~\ref{fig:4.1}. This extension significantly increases the parameter scale, enhancing the nonlinear representation capability of the draft model, enabling it to more accurately approximate the complex representation space of the target model and more effectively capture and align its feature distributions, thereby improving prediction accuracy, extending the average acceptance length, and accelerating overall inference. However, in heterogeneous architectures, when the number of Transformer blocks in the draft model exceeds a certain threshold, computational latency becomes a bottleneck. Detailed analysis and justification are provided in Section \ref{sec:4.1}.

\begin{table}
  \centering
  \resizebox{\columnwidth}{!}{
  \begin{tabular}{c >{\centering\arraybackslash}p{7.5cm}} 
    \hline
    \textbf{Category} & \textbf{Details} \\
    \hline
    Server & Intel Xeon Silver 4214R @ 2.40GHz (24 cores) \\ 
           & NVIDIA RTX 2080 SUPER (8GB VRAM) \\ 
           & PCIe Gen 3x16 \\
    Server & Intel Xeon Silver 4310 @ 2.10GHz (24 cores) \\ 
           & NVIDIA RTX 3090 (24GB VRAM) \\ 
           & PCIe Gen 4x16 \\
    PC     & Intel Core i5-9300H @ 2.40GHz (4 cores) \\ 
           & NVIDIA GTX 1050 Mobile (4GB VRAM) \\ 
           & PCIe Gen 3x8 \\
    \hline
  \end{tabular}
  }
  \caption{Hardware Configurations Employed in the Experiment.}
  \label{tab:hardware_config}
\end{table}

\begin{table*}[t]
  \centering
  \resizebox{\linewidth}{!}{%
    \begin{tabular}{cccccccccccc}
    \toprule
          &       & \multicolumn{2}{c}{MT-bench} & \multicolumn{2}{c}{HumanEval} & \multicolumn{2}{c}{GSM8K} & \multicolumn{2}{c}{Alpaca} & \multicolumn{2}{c}{Mean} \\
    \midrule
    Model & Method & Speedup & $\tau$     & Speedup & $\tau$     & Speedup & $\tau$     & Speedup & $\tau$    & Speedup & $\tau$ \\
    \midrule
    \multicolumn{12}{c}{Temperature=0} \\
    \midrule 
    \multirow{5}[2]{*}{L2 7B} 
          & Vanilla EAGLE-2 & 1.62x & 4.75 & 1.90x & 5.61 & 1.63x & \textbf{4.97} & 1.54x & \textbf{4.65} & 1.67x & \textbf{5.00} \\
          & ShearedLlama-1.3B & 1.80x & 4.75  & 2.10x & 5.48  & 1.69x & 4.41  & 1.68x & 4.47    & 1.82x & 4.78 \\
          & TinyLlama-1.1B & 1.89x & \textbf{4.89}  & 2.17x & 5.70  & 1.69x & 4.38  & 1.69x & 4.57   & 1.86x & 4.88 \\
          & EAGLE-2 & 1.99x & 3.95  & 2.32x & 4.69  & 1.99x & 4.01  & 1.93x & 3.83  & 2.06x & 4.12 \\
          & Ours & \textbf{2.25x} & 4.73 & \textbf{2.77x} & \textbf{5.90} & \textbf{2.20x} & 4.71 & \textbf{2.17x} & 4.62  & \textbf{2.35x} & 4.99 \\
    \midrule  
    \multirow{2}[2]{*}{V 13B} 
          & EAGLE-2 & 2.19 & 4.25 & 2.60 & 4.91 & 2.01x & 4.22 & 1.88x & 3.77 & 2.17x & 4.29 \\
          & Ours & \textbf{2.50x} & \textbf{5.00} &\textbf{3.19x} & \textbf{6.25} & \textbf{2.57x} & \textbf{5.12} & \textbf{2.35x} & \textbf{4.51} & \textbf{2.65x} & \textbf{5.22} \\
    \midrule
    \multicolumn{12}{c}{Temperature=1} \\
    \midrule
    \multirow{5}[2]{*}{L2 7B} 
          & Vanilla EAGLE-2 & 1.54x & 4.49 & 1.77x & 5.23 & 1.63x & \textbf{4.90} & 1.50x & \textbf{4.41} & 1.61x & \textbf{4.76} \\
          & ShearedLlama-1.3B & 1.69x & 4.37  & 1.87x & 4.83  & 1.71x & 4.52  & 1.61x & 4.21   & 1.72x & 4.48 \\
          & TinyLlama-1.1B & 1.78x & \textbf{4.53}  & 1.94x & 5.00  & 1.66x & 4.35  & 1.67x & 4.33   & 1.76x & 4.55 \\
          & EAGLE-2 & 1.88x & 3.67  & 2.14x & 4.25  & 1.96x & 3.98  & 1.81x & 3.60  & 1.95x & 3.89 \\
          & Ours & \textbf{2.12x} & 4.38 & \textbf{2.49x} & \textbf{5.34} & \textbf{2.16x} & 4.68 & \textbf{2.02x} & 4.24  & \textbf{2.20x} & 4.66 \\
    \midrule  
    \multirow{2}[2]{*}{V 13B} 
          & EAGLE-2 & 2.01x & 3.62 & 2.27x & 4.18 & 1.92x & 3.73 & 1.71x & 3.43 & 1.98 & 3.74 \\
          & Ours & \textbf{2.21x} & \textbf{4.17} & \textbf{2.62x} & \textbf{5.02} & \textbf{2.24x} & \textbf{4.43} & \textbf{2.07x} & \textbf{4.04} & \textbf{2.29} & \textbf{4.42} \\
    \bottomrule
    \end{tabular}%
    }
  \caption{A comparison of speedup ratios and average acceptance length $\tau$ for different methods on heterogeneous architectures with GeForce RTX 2080 SUPER, where L2 represents LLaMA2-Chat and V represents Vicuna.}
  \label{tab:big}%
\end{table*}

\section{Experiments}

Hardware. To validate the versatility of Dovetail in low-end hardware environments, tests were conducted in two representative scenarios: a server from the pre-large-model era and a personal computer. Detailed configurations are presented in Table~\ref{tab:hardware_config}. 


Models. In the evaluation process, LLaMA2-Chat 7B, 13B and Vicuna 13B were selected as target models to cover the performance of models at different scales.

Tasks. To comprehensively assess the performance of the models across various tasks, multiple datasets were utilized: MT-bench \cite{DBLP:conf/nips/ZhengC00WZL0LXZ23} for dialogue tasks, HumanEval \cite{chen2021evaluating} for code generation, GSM8K \cite{cobbe2021training} for mathematical reasoning, and the Alpaca dataset \cite{taori2023stanford} for instruction-following tasks.

Metrics. Given that speculative decoding inherently achieves lossless acceleration, the average acceptance length $\tau$ and the speedup ratio were chosen as the primary metrics to evaluate the acceleration performance of the target LLMs.

Training. We trained the draft model on the ShareGPT dataset, where the configuration of the draft model under the Dovetail framework involved varying the number of blocks \( M \) from 1 to 6. The training process utilized eight NVIDIA A800 80G GPUs with a batch size of 16 and employed mixed-precision training (bf16). The AdamW optimizer was used, with momentum parameters set to \( \beta_1=0.9 \) and \( \beta_2=0.95 \). The model was trained for 24 epochs, and the entire training process took approximately 1 day when \( M=6 \). To ensure a fair comparison, the EAGLE-2 model was retrained under the same conditions, providing a consistent experimental baseline.
\begin{table}[t]

\begin{center}
\resizebox{\columnwidth}{!}{%
\begin{tabular}{ccccccc}
\toprule
Method & \multicolumn{3}{c}{MT-bench} & \multicolumn{3}{c}{HumanEval} \\
\cmidrule(lr){2-4} \cmidrule(lr){5-7}
& Speedup & $\tau$ & PM & Speedup & $\tau$ & PM \\
\midrule
\multicolumn{7}{c}{L2 7B (GeForce RTX 2080 SUPER)} \\
\midrule
CPU-only & 1x(2.14t/s) & - & - & 1x(2.12t/s) & - & - \\
Offload & 0.45x & - & 7.44 & 0.45x & - & 7.44 \\
SpecExec & 2.36x & \textbf{7.43} & 7.14 & 2.98x & \textbf{10.10} & 7.32 \\
    Dovetail & \textbf{3.08x} & 4.61 & 7.40 & \textbf{3.78x} & 5.90 & 7.44 \\
\midrule
\multicolumn{7}{c}{L2 7B (GeForce RTX 3090)} \\
\midrule
CPU-only & 1x(2.35t/s) & - & - & 1x(2.34t/s) & - & - \\
Offload & 0.83x & - & 7.44 & 0.83x & - & 7.44 \\
SpecExec & 3.95x & \textbf{7.38} & 7.14 & 4.92x & \textbf{10.05} & 7.32 \\
Dovetail & \textbf{4.05x} & 4.60 & 7.40 & \textbf{4.99x} & 5.91 & 7.44 \\
\midrule
\multicolumn{7}{c}{L2 13B (GeForce RTX 3090)} \\
\midrule
CPU-only & 1x(1.20t/s) & - & - & 1x(1.22t/s) & - & - \\
SpecExec & 4.85x & \textbf{8.23} & 22.5 & 7.10x & \textbf{13.38} & 22.7 \\
Dovetail & \textbf{7.66x} & 4.53 & 21.9 & \textbf{10.14x} & 6.26 & 22.0 \\
\bottomrule
\end{tabular}%
}
\caption{Speedup ratios of different methods at temperature = 0, with PM (peak memory) in GB and tokens generated per second denoted as t/s, where L2 represents LLaMA2-Chat.}
\label{tab:method_comparison}
\end{center}
\end{table}

Parameter Settings. In the server configuration, the dynamic tree width and depth were set to 10 and 7, respectively, with 16 candidate draft tokens. In terms of model precision, the target model on the CPU employed 32-bit weights, while the draft model on the GPU used 16-bit weights. For the personal computer, the dynamic tree width and depth were adjusted to 10 and 4, respectively, with 7 candidate draft tokens. Due to the memory constraints of the personal computer, the target model on the CPU utilized 8-bit weights (obtained through PyTorch \cite{paszke2017automatic} dynamic quantization), while the draft model on the GPU continued to use 16-bit weights.

\subsection{Result}

Table~\ref{tab:big} presents the average acceptance lengths and speedup ratios of various methods across different models and temperatures. Our method achieves the highest speedup ratio in all tasks presented in the table. Specifically, the draft model optimized for heterogeneous architectures outperforms Vanilla EAGLE-2. Vanilla EAGLE-2 applies the EAGLE-2 algorithm directly on heterogeneous architectures with 60 draft tokens, whereas other methods utilize 16 draft tokens. Although reducing the number of draft tokens decreases the average acceptance length of EAGLE-2, its average speedup ratio improves from 1.67x to 2.06x.

A straightforward approach to increasing the parameter size of the draft model is to employ smaller models from the same series as the draft model. These smaller models exhibit behavioral characteristics highly consistent with the target model, significantly enhancing the average acceptance length. However, while TinyLlama-1.1B \cite{zhang2024tinyllama} and ShearedLlama-1.3B \cite{DBLP:conf/iclr/XiaGZ024} achieve average acceptance lengths of 4.88 and 4.78, respectively, the higher draft latency 
offsets the speedup gains from increased acceptance lengths in heterogeneous architectures, resulting in overall speedup performance that is only marginally better than Vanilla EAGLE-2. In contrast, Dovetail achieves an average acceptance length of 4.99 across four tasks, surpassing the smaller models in the same series while maintaining low draft latency, thus delivering the best performance across all tasks.

\begin{table}
\begin{center}
\resizebox{\columnwidth}{!}{%
\begin{tabular}{ccccc}
\toprule
Draft/Target Model & Method & Tokens/Sec & Speedup & $\tau$   \\
\midrule
- / L2 7B gptq-4bit &  Offload  & 0.45 & 0.12x & -  \\
- / L2 7B-8bit &  - / CPU-only  & 3.65 & 1x & -  \\
EAGLE-2 / L2 7B-8bit & GPU/CPU & 6.10 & 1.67x & 3.51  \\
Ours / L2 7B-8bit & GPU/CPU & \textbf{6.35} & \textbf{1.74x} & \textbf{3.78}  \\
\hline
- / V 13B-8bit &  - / CPU-only  & 1.88 & 1x & -  \\
EAGLE-2 / V 13B-8bit & GPU/CPU & 3.19 & 1.69x & 3.61  \\
Ours / V 13B-8bit & GPU/CPU & \textbf{3.36} & \textbf{1.79x} & \textbf{3.85}  \\
\bottomrule
\end{tabular}%
}
\caption{The speedup ratios of different methods were evaluated on an NVIDIA GTX 1050 using the HumanEval dataset, with the temperature set to 0. Here, L2 denotes LLaMA2-Chat, V represents Vicuna. GPU/CPU represents the heterogeneous deployment method.}
\label{tab:aba1}
\end{center}
\end{table}
As shown in Table~\ref{tab:big}, the peak memory usage of the Dovetail scheme is 2.95 GB. When the memory capacity of consumer-grade GPUs exceeds 3 GB, partial layers of the target model can be offloaded to the GPU for further acceleration. According to Table~\ref{tab:method_comparison}, in a GeForce RTX 3090 environment, 14 layers of Llama2-13B are deployed on the CPU, while the GPU hosts 26 layers of the target model and 5 layers of the draft model. Under this configuration, Dovetail achieves a speedup ratio of 10.14× on the HumanEval dataset for LLaMA2-Chat 13B, significantly outperforming SpecExec’s 7.10×. This improvement is primarily attributed to the substantial communication overhead in SpecExec’s offloading-based design, which partially offsets the benefits gained from its longer average accepted length. Additionally, both SpecExec and Dovetail use their respective optimal number of draft tokens—256 and 16. In the RTX 2080 GPU setting, for Llama2-7B, 22 layers are deployed on the CPU, while the GPU accommodates 10 layers of the target model and 5 layers of the draft model. Here, Dovetail achieves a speedup of 3.78× on HumanEval for LLaMA2-Chat 7B, surpassing SpecExec’s result of 2.98×. The numbers of draft tokens used in SpecExec and Dovetail under this configuration are 128 and 16, respectively. For further details regarding the configuration of SpecExec and the selection of draft tokens, please refer to Appendix~\ref{sec:SpecExec}.

In configurations where GPU performance significantly exceeds CPU performance, scenarios of CPU-GPU performance imbalance can be simulated. As illustrated in Table~\ref{tab:method_comparison}, with enhanced GPU performance and improved PCIe bandwidth, the performance of offloading methods improves, with the speedup ratio for LLaMA2-Chat 7B increasing from 0.45x on the GeForce RTX 2080 SUPER to 0.83x on the GeForce RTX 3090. However, this also results in a less pronounced speedup ratio improvement for Dovetail compared to SpecExec on the GeForce RTX 3090 than on the GeForce RTX 2080 SUPER. Nevertheless, Dovetail still maintains a superior speedup ratio over SpecExec, demonstrating its robust adaptability.

In personal computing environments, LLM inference is constrained by CPU memory and GPU VRAM capacity, necessitating the use of quantization techniques to reduce computational and storage overhead. It is important to emphasize that quantization algorithms directly affect model accuracy, and our primary optimization goal is to enhance the inference speed of quantized models in resource-constrained environments rather than their accuracy. As shown in Table~\ref{tab:aba1}, applying PyTorch dynamic quantization to convert the target model to 8-bit allows it to be fully loaded into the CPU memory of most personal computers. When combined with heterogeneous speculative decoding algorithms, the inference speeds of LLaMA2-Chat 7B and Vicuna 13B increase to 6.35 and 3.36 tokens per second, respectively. However, due to the limited parallel computing capability of CPUs in personal computing environments, the number of candidate tokens during the verification phase is constrained, leading to reduced average acceptance lengths and significantly lower speedup performance compared to server environments. For more details, please refer to Appendix~\ref{sec:appendix_1050}.

\subsection{Ablation Study}
In this section, we conducted an ablation study to explore the impact of DGF and multiple Transformer blocks on model performance. For more details on the tests, please refer to Appendix~\ref{sec:appendix_all}.

\subsubsection{Dynamic Gating Fusion}
To validate the effectiveness of DGF, we conducted a comparative analysis against a baseline method from EAGLE-2, in which token embeddings are linearly combined with hidden states. As shown in Table~\ref{tab:aba}, the results demonstrate that incorporating DGF significantly improves both the average acceptance length and speedup ratio on the MT-bench and HumanEval datasets. These findings highlight the ability of DGF to effectively leverage input information from multiple sources and dynamically adjust the contribution of each source, enabling more efficient and adaptive feature fusion.  
\begin{table}
\begin{center}
\resizebox{\columnwidth}{!}{%
\begin{tabular}{cccccc}
\toprule
Method & Lparameters & \multicolumn{2}{c}{MT-bench} & \multicolumn{2}{c}{HumanEval} \\
\cmidrule(lr){3-4} \cmidrule(lr){5-6}
& & Speedup & $\tau$ & Speedup & $\tau$ \\
\midrule
w/o both & 0.22B & 1.99x & 3.95 & 2.32x & 4.69 \\
w/ DGF  & 0.25B & 2.05x & 4.06 & 2.42x & 4.89 \\
w/ DGF + 1 & 0.44B & 2.13x & 4.31 & 2.62x & 5.38 \\
w/ DGF + 2 & 0.63B & 2.21x & 4.53 & 2.72x & 5.65 \\
w/ DGF + 3 & 0.81B & 2.23x & 4.62 & 2.74x & 5.82 \\
w/ DGF + 4 & 1.00B & 2.25x & 4.73 & \textbf{2.77x} & 5.90 \\
w/ DGF + 5 & 1.19B & \textbf{2.26x} & \textbf{4.83} & 2.75x & \textbf{5.98} \\
\bottomrule
\end{tabular}%
}
\caption{Ablation experiment results on a heterogeneous architecture using GeForce RTX 2080 SUPER, with the temperature set to 0 for LLaMA2-Chat-7B. Lparameters denotes the model's learnable parameters. w/o both indicates using only one layer, w/ DGF indicates using one layer with DGF, w/ DGF + m indicates using w/ DGF with an additional m Transformer blocks.}
\label{tab:aba}
\end{center}
\end{table}
\subsubsection{Multiple Transformer Blocks}\label{sec:4.1}

To evaluate the impact of the draft model's parameter scale on inference speed, we gradually increased the number of Transformer blocks in the draft model from 1 to 6. As shown in Table~\ref{tab:aba}, increasing the number of Transformer blocks from 1 to 5 led to a gradual improvement in prediction accuracy, which in turn resulted in a corresponding increase in average acceptance length and a steady rise in the speedup ratio. This indicates that increasing the number of Transformer blocks enables the model to capture more complex features, thereby aligning the draft model's feature distribution more closely with that of the target model. However, when the number of Transformer blocks reached 6, while both the average acceptance length and speedup ratio improved significantly on the MT-bench dataset, the speedup ratio on the HumanEval dataset slightly decreased despite a marked improvement in average acceptance length. This phenomenon can be attributed to the fact that, at this stage, the inference time during the draft phase becomes the primary bottleneck. The additional parameters significantly increase the draft computation time, which offsets the acceleration benefits gained from the improved average acceptance length.


\section{Related work}

\subsection{Heterogeneous Architecture}


Transformer \cite{vaswani2017attention} and its variants have emerged as the dominant architecture for LLMs. However, the increasing scale of these models has led to inference speed being constrained by the memory capacity of accelerators. To address this challenge, researchers have proposed various compression techniques, such as quantization \cite{hubara2018quantized,xiao2023smoothquant,frantar2022gptq,DBLP:conf/icml/LiuYJZXBC024,DBLP:conf/emnlp/YuanLZCLWLJCX0H24}, pruning \cite{gale2019state,liu2023deja}, and knowledge distillation \cite{sanh2019distilbert,DBLP:conf/acl/TuPWG20,DBLP:conf/acl/Wen0DM23}. However, these methods often come at the cost of degraded generation quality. To achieve lossless inference, offloading stores parameters exceeding GPU capacity in CPU memory and dynamically loads them to the GPU when needed. However, 99.5\% of the time in single-batch inference is spent on data transfer \cite{DBLP:conf/sosp/SongMX024}, significantly increasing latency. Partial offloading \cite{llama.cpp} directly computes the excess parameters on the CPU and transfers intermediate results to the GPU for further processing, but its performance remains constrained by the computational capabilities of the CPU and the memory capacity of the GPU. Future research aims to combine the characteristics of models with the specific advantages of heterogeneous architectures to achieve more efficient inference acceleration. 

In heterogeneous architectures, the presence of accelerators allows for leveraging the advantages of multiple computational resources for LLM inference. Model compression techniques \cite{zhang202570} typically focus on fully utilizing accelerator performance, often with limited consideration of output quality. In contrast, offloading and partial offloading strategies combine the performance of accelerators with the memory and computational capabilities of CPUs to achieve lossless output quality, although their acceleration efficiency is generally suboptimal. To address this issue, PowerInfer \cite{DBLP:conf/sosp/SongMX024} leverages the locality characteristics of LLM inference by predicting hot neurons to be computed on the GPU, while delegating cold neurons to the CPU. This approach effectively utilizes the advantages of heterogeneous architectures to significantly improve inference speed. Similarly, KTransformers \cite{ktransformers} focuses on sparse Mixture of Experts (MoE) models, employing a heterogeneous computing strategy: non-shared components (sparse MoE matrices) are placed on the CPU to conserve GPU memory, while shared dense components are computed on the GPU. This method maximizes hardware resource utilization through heterogeneous computing, enabling efficient inference in resource-constrained environments.

\subsection{Speculative Decoding}

Speculative decoding is an emerging lossless acceleration method based on the draft-then-verify paradigm \cite{DBLP:conf/acl/XiaYDW00L0S24}, which can be outlined from the following three aspects.

\subsubsection{Obtaining Draft Tokens}
For certain target models \cite{touvron2023llama,yang2024qwen2}, smaller models from the same series can be directly used as draft models \cite{DBLP:conf/icml/LeviathanKM23} without requiring additional training or modification. When small models from the same series are unavailable, the draft model must be trained from scratch, or draft models or draft tokens can be derived from the target model. 
Draft models can be obtained from target models using knowledge distillation \cite{zhou2023distillspec} or quantization \cite{miao2023specinfer}, or by incorporating early exit mechanisms \cite{DBLP:conf/aaai/ZengHDZC24} and layer-skipping techniques \cite{DBLP:conf/acl/Zhang00S0CM24} to conclude the inference process earlier, thus generating draft tokens.Additionally, non-autoregressive or autoregressive prediction heads \cite{DBLP:conf/icml/CaiLGPLCD24,DBLP:conf/icml/LiW0024} can be incorporated into the target model to generate draft tokens. A draft model can also be composed of multiple smaller models, leveraging a staged \cite{spector2023accelerating} or cascaded approach\cite{DBLP:conf/nips/0003YLSC024} to generate draft tokens.
\subsubsection{Organizing Draft Tokens}
In early studies \cite{DBLP:conf/icml/LeviathanKM23,chen2023accelerating}, the draft model sampled only one draft token per step and used a chain structure. To increase average acceptance length, later studies \cite{miao2023specinfer,DBLP:conf/icml/CaiLGPLCD24} proposed sampling multiple draft tokens per step and organizing them in a predefined tree structure. However, static tree structures do not consider contextual information. Studies \cite{DBLP:conf/nips/SvirschevskiMCC24,DBLP:conf/emnlp/LiW0024}have suggested dynamically constructing a draft tree based on the cumulative confidence of tokens in their context.

\subsubsection{Verifying Draft Tokens}
 Early studies \cite{stern2019insertion,DBLP:conf/emnlp/Xia0WCWS23} strictly required that draft tokens match the greedy decoding output of the target model exactly. Later, speculative sampling \cite{DBLP:conf/icml/LeviathanKM23,chen2023accelerating} adopted nucleus sampling and theoretically demonstrated that this criterion preserves the same output distribution as the target LLM, also achieving lossless acceleration. To further enhance acceleration, some studies \cite{DBLP:conf/emnlp/Xia0WCWS23,kim2024speculative} have proposed moderately relaxing the verification criteria. Judge decoding \cite{DBLP:conf/iclr/BachmannAPGSDST25} can determine whether to accept a draft token directly based on its token embedding, without relying on logits.

\section{Conclusion}
This paper proposes a lossless acceleration method named Dovetail, which employs speculative decoding to optimize the inference efficiency of target models under resource-constrained conditions. Tailored for low-end hardware characteristics, Dovetail reduces the number of draft tokens, thereby linearly decreasing the latency of parallel verification, and utilizes DGF to efficiently integrate multi-source information. Additionally, by increasing the parameter size of the draft model, it enhances prediction accuracy, achieving a higher speedup ratio. Experimental results demonstrate that Dovetail outperforms existing lossless acceleration methods across multiple datasets and achieves the highest speedup ratio in all benchmark tests.


\section*{Limitations}

Although the proposed method has achieved relatively superior performance, achieving optimal inference speed in resource-constrained environments still needs more effort. Due to the limitations of CPU parallelism, inference methods face challenges when dealing with long text scenarios because the delay in the pre-filling stage is relatively large. This is a task that needs to be addressed in the future.

\section*{Acknowledgments}
This work is sponsored in part by the National Natural Science Foundation of China under Grant No. 62025208 and 62421002.

\bibliography{acl_latex}

\appendix

\section{Dovetail}
\begin{figure}[t]
  \centering
  \includegraphics[width=0.6\columnwidth]{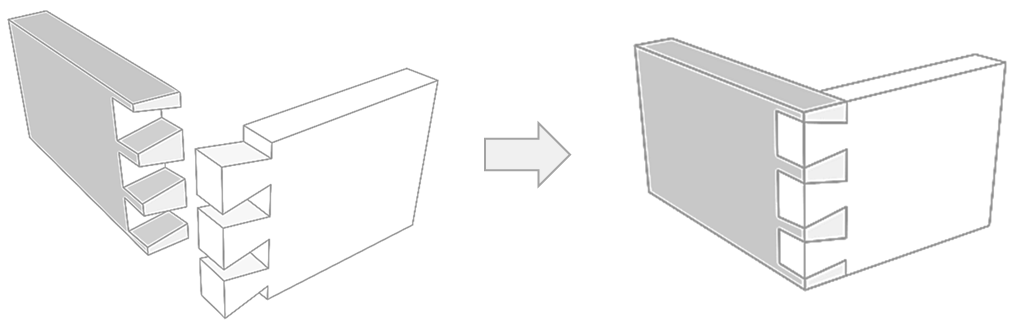}
    \caption{A depiction of the Dovetail joint in Chinese carpentry, which also inspired the name of our method. It represents that the seamless integration of this heterogeneous architecture.}
    \label{fig:4.233333}
\end{figure}

Figure~\ref{fig:4.233333} is an illustrative description of Dovetail.

\section{Analysis of Dovetail Effectiveness}
\label{sec:appendix}

Given the sequence length \( S \), batch size \( B \), a target model consisting of \( m \) Transformer blocks, hidden dimension \( H \), and the number of candidate draft tokens \( \gamma \), the average decoding latency per token based on the theoretical formula of MagicDec is defined as:  
\begin{equation}  
T_{\mathit{Avg}}^{\mathit{SD}} = \frac{\gamma \cdot T_D + T_V(\gamma)}{\Omega(\gamma, \alpha)}
\label{eq:avg-latency}  
\end{equation}  
where \( \alpha \) is the acceptance rate, \( \Omega(\gamma, \alpha) \) is the number of accepted tokens in a single parallel verification, \( T_D \) is the time for the draft model to decode a single token, and \( T_V(\gamma) \) is the time for the target model to verify \( \gamma \) tokens in parallel.  

The offloading method employs a strategy of overlapping computation and data loading to optimize efficiency, with the latency per token denoted as \( T_{\mathit{Offload}} \). To ensure the advantage of the heterogeneous speculative decoding method, the following condition must be satisfied:  
\begin{equation}  
T_{\mathit{Avg}}^{\mathit{SD}} < T_{\mathit{Offload}}
\label{eq:condition}  
\end{equation}  
i.e., the average latency per token of heterogeneous speculative decoding must be lower than that of the offloading method.  

In a heterogeneous architecture, the parallel verification time \( T_V(\gamma) \) of the target model on the main processor can be decomposed into computation time \( T_c \) and memory access time \( T_p \):  
\begin{equation}  
T_c = \frac{F}{P_{\mathit{c}} \cdot E_{\mathit{c}}}, \quad T_p = \frac{P}{B_{\mathit{m}} \cdot E_{\mathit{m}}},  
\end{equation}  
where \( F \) is the computational cost of a single Transformer block, \( P \) is the parameter size, \( P_c \) is the peak computational performance of the main processor, \( E_c \) is the computational efficiency, \( B_m \) is the memory bandwidth, and \( E_m \) is the memory utilization efficiency.  

Considering the partial overlap between computation time and memory access time, the total time is expressed as:  
\begin{equation}  
T_V(\gamma) = \max(T_c, T_p) + \beta \cdot \min(T_c, T_p)
\end{equation}  
where \( \beta \) (\( 0 \leq \beta \leq 1 \)) is the overlap coefficient, with \( \beta = 1 \) indicating no overlap and \( \beta = 0 \) indicating complete overlap.  

Equation~\ref{eq:condition} can be further updated as:  
\begin{equation}  
\max(T_c, T_p) + \beta \cdot \min(T_c, T_p) < \Omega(\gamma, \alpha) \cdot \frac{P}{B_{\mathit{p}} \cdot E_{\mathit{p}}}
\end{equation}  
where \( T_c \) and \( T_p \) are the computation time and memory access time of the target model on the main processor, \( \beta \) (\( 0 \leq \beta \leq 1 \)) is the overlap coefficient, \( B_p \) is the PCIe bandwidth, and \( E_p \) is the PCIe transmission efficiency.  

In summary, heterogeneous speculative decoding achieves higher efficiency when the performance of the main processor and the accelerator is relatively balanced. However, in scenarios with severe hardware resource imbalance (e.g., significantly limited computational performance of the main processor and high PCIe bandwidth), heterogeneous speculative decoding may not be the optimal choice.

\section{Additional Implementation Details}
\label{sec:appendix_all}

\subsection{Ablation experiments on RTX 2080}
\label{sec:appendix_2080}

\begin{table}
\begin{center}
\begin{tabular}{ccc}
\toprule
Method & Speedup & $\tau$   \\
\midrule
w/o both  & 2.60x & 4.91 \\
w/ DGF   & 2.69x & 4.99 \\
w/ DGF + 1  & 2.97x & 5.71 \\
w/ DGF + 2   & 3.04x & 5.96 \\
w/ DGF + 3  & 3.08x & 6.03 \\
w/ DGF + 4 & \textbf{3.19x} & 6.25 \\
w/ DGF + 5   & 3.13x & \textbf{6.26} \\
\bottomrule
\end{tabular}%
\caption{The ablation experiment results of Vicuna 13B on a heterogeneous architecture using GeForce RTX 2080 SUPER, with the temperature set to 0 and the test dataset being HumanEval. “w/o both” denotes using only a single layer; “w/ DGF” indicates using a single layer with DGF; and “w/ DGF + m” represents adding m additional Transformer blocks on the basis of “w/ DGF”.}
\label{tab:aba5}
\end{center}
\end{table}


When running the Vicuna 13B on a GeForce RTX 2080, the parameter scale of the draft model also significantly impacts the inference speed, as shown in Table~\ref{tab:aba5}. As the number of Transformer blocks in the draft model increases from 1 to 5, the prediction accuracy progressively improves, driving a corresponding increase in the average acceptance length, while the speedup ratio steadily rises. However, when the number of Transformer blocks increases to 6, although the average acceptance length shows a notable improvement, the speedup ratio experiences a slight decline. This phenomenon aligns with the observations made on the 7B model.

\subsection{Ablation experiments on GTX 1050}
\label{sec:appendix_1050}


\begin{table}
\begin{center}
\begin{tabular}{cccc}
\toprule
 Method & Tokens/Sec & Speedup & $\tau$   \\
\midrule
w/o both & 3.19 & 1.69x & 3.61  \\
w/DGF & 3.36 & 1.79x & 3.85  \\

w/DGF+1 & 2.85 & 1.52x & 4.09  \\
\bottomrule
\end{tabular}%

\caption{The ablation study results of the Vicuna 13B  on HumanEval, conducted on a heterogeneous architecture with NVIDIA GTX 1050, where the temperature is set to 0.}
\label{tab:aba2}
\end{center}
\end{table}

We conducted ablation experiments on the Vicuna 13B on a platform equipped with an NVIDIA GTX 1050 to investigate the impact of the DGF module and multiple Transformer blocks on model performance. The experimental results are presented in Table~\ref{tab:aba2}. Upon integrating the DGF module into EAGLE-2, both the speedup ratio and the average acceptance length of the model exhibited improvements. However, when an additional Transformer block was introduced beyond this configuration, while the average acceptance length continued to increase, the speedup ratio experienced a decline. The primary reason for this phenomenon, as shown in Table~\ref{tab:aba3}, lies in the precision discrepancy between the CPU and GPU: the CPU employed int8 quantization, whereas the GPU utilized fp16 precision for computations. This precision mismatch resulted in an insufficient time difference between the drafting phase and the parallel verification phase to accommodate the inclusion of an extra Transformer block. As shown in Table~\ref{tab:aba2}, further increasing the number of Transformer blocks prolonged the drafting time, thereby diminishing the overall acceleration effect. Consequently, to achieve performance akin to that of a RTX 2080 on a device such as the GTX 1050—specifically, to further enhance the speedup ratio by incorporating additional Transformer blocks—it is advisable to apply int8 quantization to the drafting model on the GPU. This approach would amplify the time difference between the drafting phase and the parallel verification phase, thereby enabling the integration of multiple additional blocks.

\begin{table}[t]
\begin{center}
\begin{tabular}{ccc}
\toprule
  & Draft Stage  & Verify Stage     \\
\midrule
Precision & 16-bit  & 8-bit   \\
Processors  & GTX 1050   &  i5-9300H   \\
Time & 0.31 Sec  & 0.57 Sec   \\
\bottomrule
\end{tabular}%
\caption{On a personal laptop, statistics were gathered for large-model inference using w/DGF, with a focus on the average time taken for a single drafting phase and the average time taken for a single parallel verification phase.}
\label{tab:aba3}
\end{center}
\end{table}


\subsection{Experiments Related to SpecExec}
\label{sec:SpecExec}

\begin{table}[t]
\centering
\resizebox{0.48\textwidth}{!}{%
\begin{tabular}{cccccc}
\toprule
Draft Tokens & \multicolumn{2}{c}{MT-bench} & \multicolumn{2}{c}{HumanEval} \\
\cmidrule(lr){2-3} \cmidrule(lr){4-5}
& Speedup & $\tau$ & Speedup & $\tau$ \\
\midrule
16 & 1.86x & 4.85 & 2.25x & 6.32 \\
128 & 2.36x & 7.43 & 2.98x & 10.10 \\
\bottomrule
\end{tabular}%
}
\caption{Performance comparison of SpecExec with different draft token counts on MT-bench and HumanEval benchmarks.}
\label{tab:specexec_draft_tokens}
\end{table}

The default configuration of SpecExec uses the target model Llama2-Chat-7B and the draft model TinyLlama-1.1B, optimized based on Offload, where both the target and draft models are executed on the GPU and the CPU is only used to store parameters. In Table~\ref{tab:method_comparison} of the paper, SpecExec uses the optimal draft count, for example, in a GPU 2080, the draft count is 128 rather than 16. Using 16 would significantly reduce performance, as shown in the table~\ref{tab:specexec_draft_tokens}, and would not provide a fair comparison with Dovetail. When the target model is Llama-2 7B and the GPU memory is 4GB, SpecExec encounters an out-of-memory error, while Dovetail runs normally.

\end{document}